\documentclass[conference]{IEEEtran}
\usepackage{times}

\usepackage[numbers]{natbib}
\usepackage{multicol}
\usepackage[bookmarks=true]{hyperref}

\usepackage{amsmath}
\usepackage{amssymb}
\usepackage{amsfonts}

\usepackage{tabu}
\usepackage{booktabs} % for professional tables
\usepackage{multirow}
\usepackage{adjustbox}
\usepackage{siunitx}

\usepackage{xcolor}
\usepackage{xspace}
\usepackage{ifthen}

\usepackage{graphicx}
\usepackage[font=small]{caption}
\usepackage{tikz}
\usetikzlibrary{positioning}
\usepackage{subfigure}

\let\labelindent\relax %needed for IEEE conflict with enumitem
\usepackage{enumitem} % for \begin{description}[align=left]

\usepackage{autolabtools}

\newcommand{\algabbr}{SSVTP}

\definecolor{dblue}{rgb}{0.122, 0.435, 0.698}
\definecolor{lblue}{rgb}{0.03, 0.1, 0.17}
\usepackage[framemethod=tikz]{mdframed}
\newmdenv[innerlinewidth=0.5pt, roundcorner=4pt,linecolor=dblue,innerleftmargin=6pt,
innerrightmargin=6pt,innertopmargin=6pt,innerbottommargin=6pt,fill=lblue]{mybox}

\begin{document}

\title{
Self-Supervised Visuo-Tactile Pretraining\\to Locate and Follow Garment Features
}

\author{Justin Kerr\authorrefmark{1}\authorrefmark{2}, Huang Huang\authorrefmark{1}\authorrefmark{2},  Albert Wilcox\authorrefmark{2},  
 Ryan Hoque\authorrefmark{2}, \\ Jeffrey Ichnowski\authorrefmark{2}\authorrefmark{5}, Roberto Calandra\authorrefmark{3}\authorrefmark{4}, and Ken Goldberg\authorrefmark{2}% <-this % stops a space
 \\
 \authorblockA{\authorrefmark{1}Equal contribution}
 \authorblockA{\authorrefmark{2}The AUTOLab at UC Berkeley\\
 \authorrefmark{3}Learning, Adaptive Systems, and Robotics (LASR) Lab, TU Dresden\\
 \authorrefmark{4}The Centre for Tactile Internet with Human-in-the-Loop (CeTI)\\
 \authorrefmark{5}Carnegie Mellon University}
 
}

\maketitle

\begin{abstract}
Humans make extensive use of vision and touch as complementary senses, with vision providing global information about the scene and touch measuring local information during manipulation without suffering from occlusions. While prior work demonstrates the efficacy of tactile sensing for precise manipulation of deformables, they typically rely on supervised, human-labeled datasets. We propose Self-Supervised Visuo-Tactile Pretraining (\algabbr), a framework for learning multi-task visuo-tactile representations in a self-supervised manner through cross-modal supervision. We design a mechanism that enables a robot to autonomously collect precisely spatially-aligned visual and tactile image pairs, then train visual and tactile encoders to embed these pairs into a shared latent space using cross-modal contrastive loss. We apply this latent space to downstream perception and control of deformable garments on flat surfaces, and evaluate the flexibility of the learned representations \textit{without fine-tuning} on 5 tasks: feature classification, contact localization, anomaly detection, feature search from a visual query (e.g., garment feature localization under occlusion), and edge following along cloth edges. 
The pretrained representations achieve a 73-100\% success rate on these 5 tasks.

\end{abstract}
\IEEEpeerreviewmaketitle

\section{Introduction} \label{sec:introduction}
Despite significant recent progress in applications of tactile sensing and deep learning to deformable object manipulation \cite{alberto2022,seitasingulate,luo2018vitac,yuan2018}, many techniques still rely on human-labeled data for training these systems.
If robots could leverage self-supervised data instead, scaling learned methods would be far more practical as robots could autonomously collect their own data. How can we achieve self-supervision for training flexible tactile sensing systems? Natural intelligence suggests one answer: multi-sensory fusion of vision and touch, learned through self-supervision, plays a critical role in early human cognitive development of manipulation capabilities \cite{lessonsfrombabies,newlysighted2011}. 
In this paper we explore how cross-modal learning from paired visual and tactile images collected in a self-supervised manner can be used to locate and follow garment features like seams, buttons, zippers, and edges.

Prior work on visuo-tactile representation learning typically relies on significant human supervision for data collection \cite{yuan2017connecting,luo2018vitac, takahashi2019deep,yang2022touch} or learns task-specific representations \cite{Calandra_2018,bauza2019mapping,Lee2019MakingSO,lin2019learning,Pecyna2022VisualTactileMF}. In this work, we propose a self-supervised data collection pipeline that uses a robot with a custom end effector to autonomously collect visual and tactile images that are \textit{spatially aligned}. This allows the learned representation to capture subtle local details in appearance and texture, making it useful for cross-modal zero-shot queries for downstream tasks. We use the paired data to train a visuo-tactile latent space with a cross-modal contrastive loss, a technique shown to be highly effective for vision and language representations like CLIP \cite{Radford2021LearningTV}. We decouple texture from rotation by training the latent space to be rotation-invariant and training a separate network to predict the relative rotation between a visual image and tactile image. We refer to this method as Self-Supervised Visuo-Tactile Pretraining (\algabbr) because it pretrains representations \textit{once} on self-supervised data, then deploys on multiple tasks.

\begin{figure}
    \centering
    \includegraphics[width=\columnwidth]{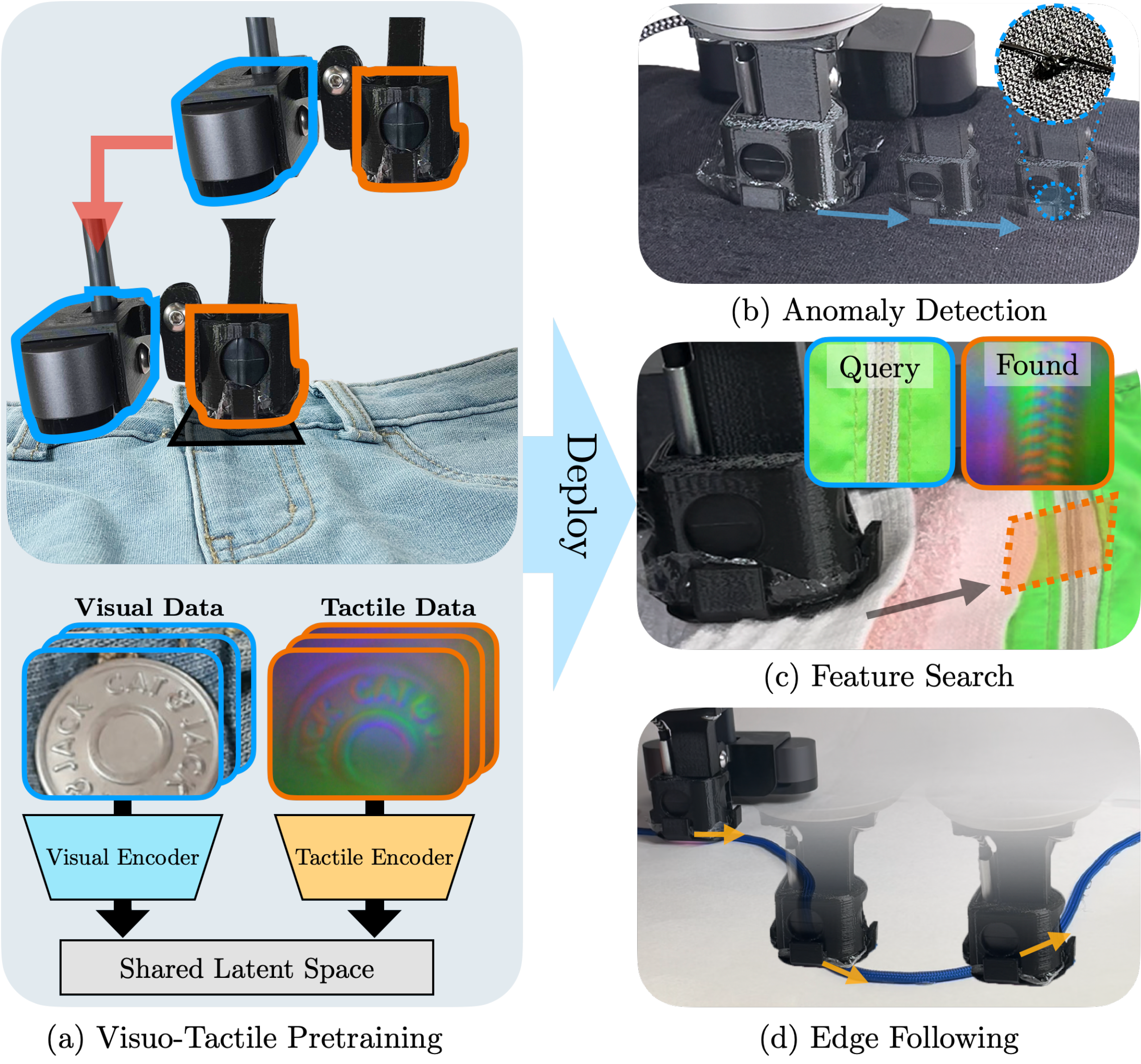}
    \caption{\textbf{\algabbr{} Overview.} (a) We design a self-supervised framework described in Section~\ref{ssec:data_col} to collect 4500 spatially aligned visual and tactile images, and use this dataset to learn a shared visuo-tactile latent space $\mathcal{Z}$ as described in Section~\ref{ssec:encoders}. We apply this latent space without fine-tuning for 3 active sliding perception tasks: Anomaly Detection (b), Feature Search (c), and Edge Following (d).}
    \label{fig:splash}
    \vspace{-20pt}
\end{figure}
We show that \algabbr{} can be used \textit{without fine-tuning} for 3 active sliding perception tasks and 2 passive perception tasks on a range of garment seams, buttons, zippers, strings, and edges.
An overview of the proposed approach is presented in Figure~\ref{fig:splash}. This paper makes the following contributions:
\begin{enumerate}[nosep,leftmargin=1.5em]
\item A novel framework and hardware design for self-supervised real-world data collection of spatially aligned visuo-tactile image pairs on deformable surfaces containing garment features.
\item A novel representation learning method decoupling texture and rotation by training a rotation-invariant latent space using contrastive loss and a separate network to predict differences in rotation.
\item Experiments using the learned representations for 3 sliding perception tasks (tactile anomaly detection, vision-guided search, and active tactile servoing) and 2 passive tasks (localization and classification).
\item A publicly available dataset with 4500 samples of real-world, spatially aligned visual and tactile images of deformable surfaces (in contrast to prior large datasets without such alignment, or on rigid objects~\cite{yuan2017connecting,touchingnerf}).

\end{enumerate}
\section{Related Work} \label{sec:relatedwork}
\subsection{Tactile Sensing for Robotics}
Tactile sensing is a longstanding challenge in robotics \cite{wearablehaptics}; in 1984 \citet{Goldberg1984ActiveTA} explored active touch sensing using an elasto-resistive foam on a ``finger."
% Tactile sensing is a longstanding challenge in robotics, with early work on force-torque sensors \cite{9591615}. 
Since then, various tactile sensors with different sensing techniques, resolutions, and form factors have been studied such as force-torque sensors \cite{9591615, Cutkosky2008}, haptic devices \cite{wearablehaptics, deepxpalm}, and soft ``skin" that mimics human skin \cite{bioeskin}. 
New sensors widely used in the past few years are the DIGIT \cite{Lambeta_2020} and the GelSight \cite{Dong_2017}. These sensors illuminate a flexible membrane with colored LED light from different directions and observe the back of the membrane with a camera. When the membrane is pressed into surfaces, indentations are illuminated to form an image. 

Tactile sensing is of renewed interest for robotic learning and manipulation. Reinforcement learning (RL) policies trained with tactile information can generalize to new object geometries for insertion tasks \cite{dong2021tactile}. \citet{lu2022curiosity} develop a tactile based SLAM method to reconstruct object geometries. Object pose estimation also benefits from tactile information~\cite{bauza2022tac2pose} and can be applied to manipulation tasks such as bottle opening~\cite{kelestemur2022tactile}. Tactile servoing, the use of tactile sensing to control the robot along local features, is another active area of research with both learning-based and classical approaches \cite{She2020CableMW, servoing1, servoing2, servoing3}. Tactile information generation in simulation has been studied \cite{gomes2021generation} and can be used with RL for zero-shot sim-to-real transfer to manipulation tasks~\citep{church2022tactile}. ~\citet{alberto2022, midastouch} incorporate tactile sliding for cloth unfolding and touch localization on known object geometries, respectively. In this work, though we implement sliding tactile primitives on cloth, our method is purely self-supervised in contrast to \cite{alberto2022,She2020CableMW}. Prior work on tactile localization uses contact depth estimates trained in simulation \cite{midastouch,lu2022curiosity}, which transfers well to rigid objects but cannot capture the high-frequency texture information required for perceiving fine garment features. In addition, simulating realistic tactile texture is currently prohibitively expensive for complex objects like deformable surfaces \cite{Seita2020DeepIL}. This work learns a visuo-tactile association from real data and applies it to robot tasks on garments without fine-tuning.

\subsection{Visuo-Tactile Cross-Modal Learning}
Learning a representation from tactile and visual inputs has been studied in the context of lifelong learning \cite{Zheng2021lifelong} and synthesizing sensory inputs from their counterparts ~\cite{takahashi2019deep,li2019connecting}. ~\citet{takahashi2019deep} use a low-resolution 16$\times$16 grid of pressure sensing ``taxels," and train an encoder-decoder structure to predict taxel values based on visual images. ~\citet{li2019connecting, lee2019touching} use video frames to predict outputs of a GelSight tactile sensor given a visual image of a robot touching a surface and vice versa using a conditional Generative Adversarial Network (cGAN) architecture. Recent work uses a cGAN conditioned on the output of a neural radiance field \cite{nerf} to synthesize tactile images \cite{touchingnerf}. In contrast, the goal of this paper is not to synthesize tactile images from vision input but instead learn a shared representation from real vision and tactile sensor data for application to robot perception and control.

~\citet{yuan2017connecting} learn an embedding space by jointly training CNNs for three modalities: RGB images, depth images, and tactile images; they use the distance in the embedding space to match among modalities and apply their methods to infer tactile properties of fabric from vision. Similarly, ~\citet{luo2018vitac} propose a fusion method to associate tactile and vision data of fabrics by applying maximum covariance analysis for the latent vectors. ~\citet{Zambelli2020LearningRT} study several approaches for cross-modal learning of vision and touch from self-supervised data. However, these papers do not explicitly apply their representations to robot control tasks.  %Compared to these works, we do not use the depth images and our dataset is collected automatically using a robot. 
Additionally, the visual and tactile images we use in this work are spatially aligned while most visual inputs in prior work have a field of view orders of magnitude larger than the tactile images, making it difficult for networks to learn or actively perceive subtleties in local texture important for identifying garment features.
\section{Problem Statement} \label{sec:problemstatement}

\begin{figure}[t!]
\vspace{1pt}
    \centering
    \subfigure[Capture Visual Image]{\includegraphics[width=.47\columnwidth]{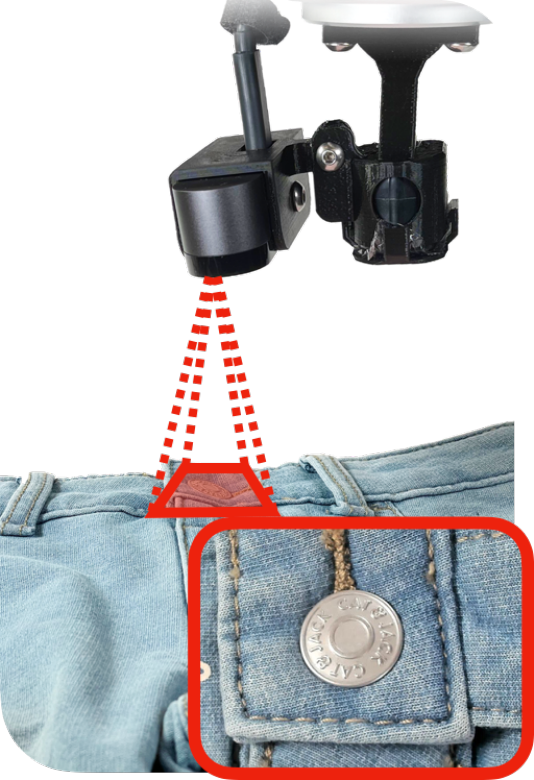}}
    \subfigure[Capture Tactile Image]{\includegraphics[width=.47\columnwidth]{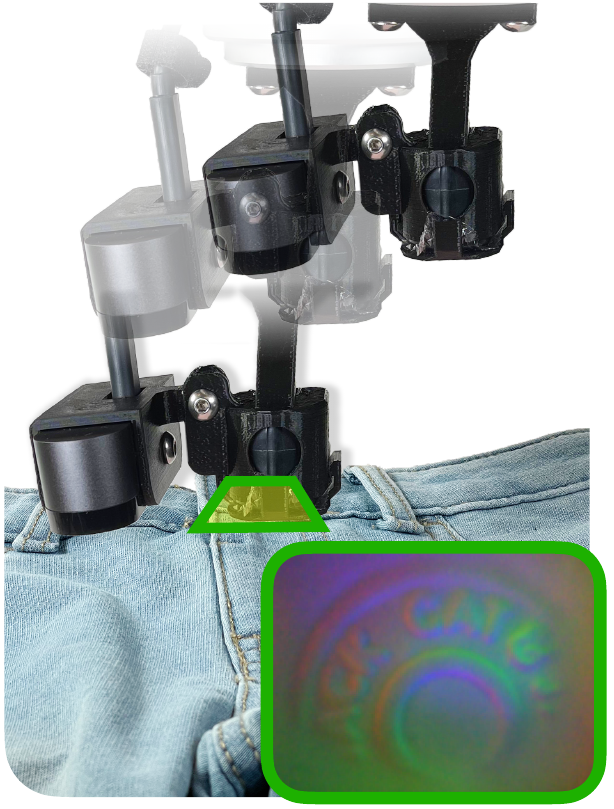}}
    \subfigure[Example Deformable Surface]{\includegraphics[width=.95\columnwidth]{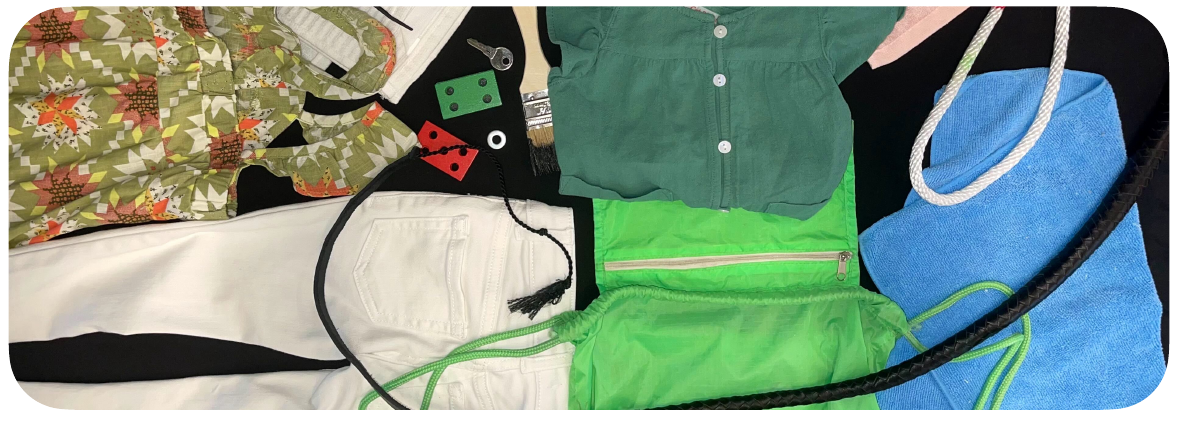}}
    \caption{\textbf{Self-Supervised Visuo-Tactile Data Collection:} An overview of the data collection pipeline. (a) For each sample the robot first uses an RGB camera to take a closely cropped image of the texture. (b) The robot then adjusts its end effector to take a tactile reading at the same location. (c) We collect data on 10 deformable surface environments, one of which is shown here.}
    \label{fig:data_col}
    \vspace{-15pt}
\end{figure}

We consider the problem of learning a cross-modal representation between vision and tactile data that can be used for localizing and tracking garment features without fine-tuning. 
We define \textit{deformable surfaces} as 2.5-dimensional environments of area up to $50$cm$\times$$50$cm and height up to \SI{1}cm consisting of layers of textured deformable objects. These deformable layers can include \textit{garment features} with varying textures like edges, corners, seams and embedded rigid objects like buttons, zippers, and rivets.
Figure~\ref{fig:data_col} provides an example of a deformable surface environment. 
Given a robot with a tactile sensor and RGB camera rigidly attached to each other and to a robot end effector and a set of deformable surfaces,
the goal is to output a function $f: O_\mathrm{vis} \times O_\mathrm{tac} \rightarrow \mathcal{Z}$ where $O_\mathrm{vis} = \mathbb{R}^{H\times W}$ and $O_\mathrm{tac} = \mathbb{R}^{H\times W \times 3}$ are the input spaces of grayscale visual images and tactile images respectively, and $\mathcal{Z}$ is an output representation space of much lower dimension than inputs.

We make the following assumptions:

\begin{enumerate}[nosep,leftmargin=1.5em]
\item Objects in deformable surface environments are 1-dimensional or 2-dimensional deformable objects (e.g., fabric or smal rigid features like buttons).
\item Deformable surface environments are arranged on a flat tabletop in stable poses.
\item Surfaces in downstream tasks are within the training distribution (i.e. similar deformable objects).
\end{enumerate}

 During evaluation, we keep $f$ constant across tasks and evaluate by measuring the performance of policies using $f$ on five tasks involving locating and follow garment features: 
 \begin{enumerate}[nosep,leftmargin=1.5em]

     \item \textit{Anomaly Detection} (Active Sliding): While collecting a stream of tactile readings during a sliding motion (see Section~\ref{ssec:manip_policies}), the task objective is to localize where a tactile anomaly occurs (e.g., a knot in an otherwise straight thread). A trial is successful if the robot stops while touching the anomaly.
     \item \textit{Feature Search} (Active Sliding):  
     Given a visual query image, the task objective is to search the workspace with the tactile sensor to find a match. A trial is successful if the robot stops while touching the feature contained in the query image.
     \item \textit{Edge Following} (Active Sliding): Given a reference visual image of a uniform directional feature (e.g., cable or towel edge), and a line in the image frame specifying which direction to slide in, the objective is to servo the tactile sensor along the edge without losing contact. We measure the distance followed without losing continuous contact between the sensor and edge. 
     \item \textit{Feature Classification} (Passive): Given a tactile query image and a set of canonical visual images (one image per class), the task objective is to determine which class the tactile image corresponds to.
     \item \textit{Contact Localization} (Passive): Given a tactile query image and a visual observation of the entire 2D workspace, the task objective is to localize in the visual image where the tactile reading was taken.
 \end{enumerate}
 
\begin{figure}[t!]
\vspace{1pt}
    \centering
    \subfigure[CAD Model]{
        \includegraphics[width=0.4\columnwidth]{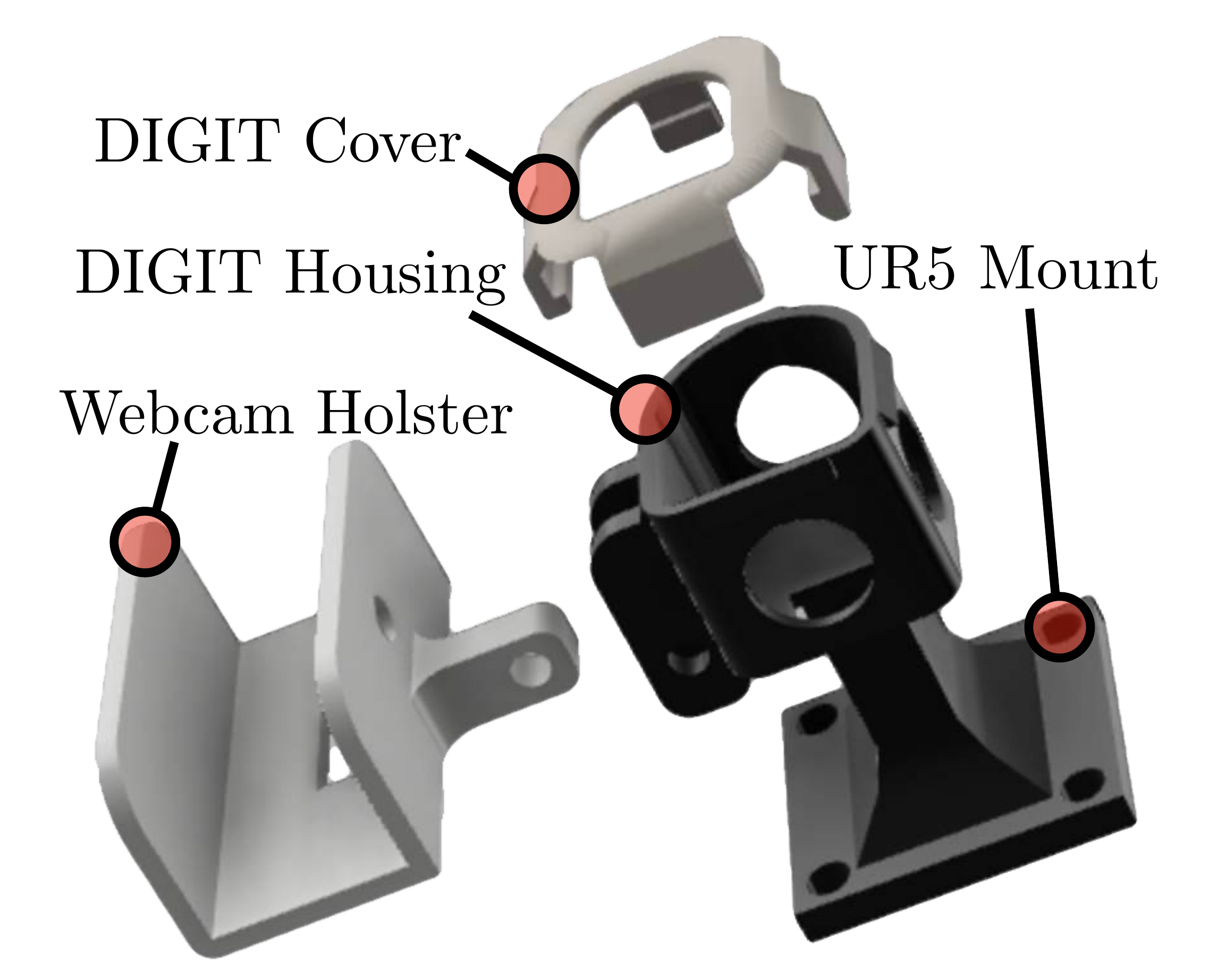}
    }
    \subfigure[3D Printed Hardware]{
        \includegraphics[width=0.4\columnwidth]{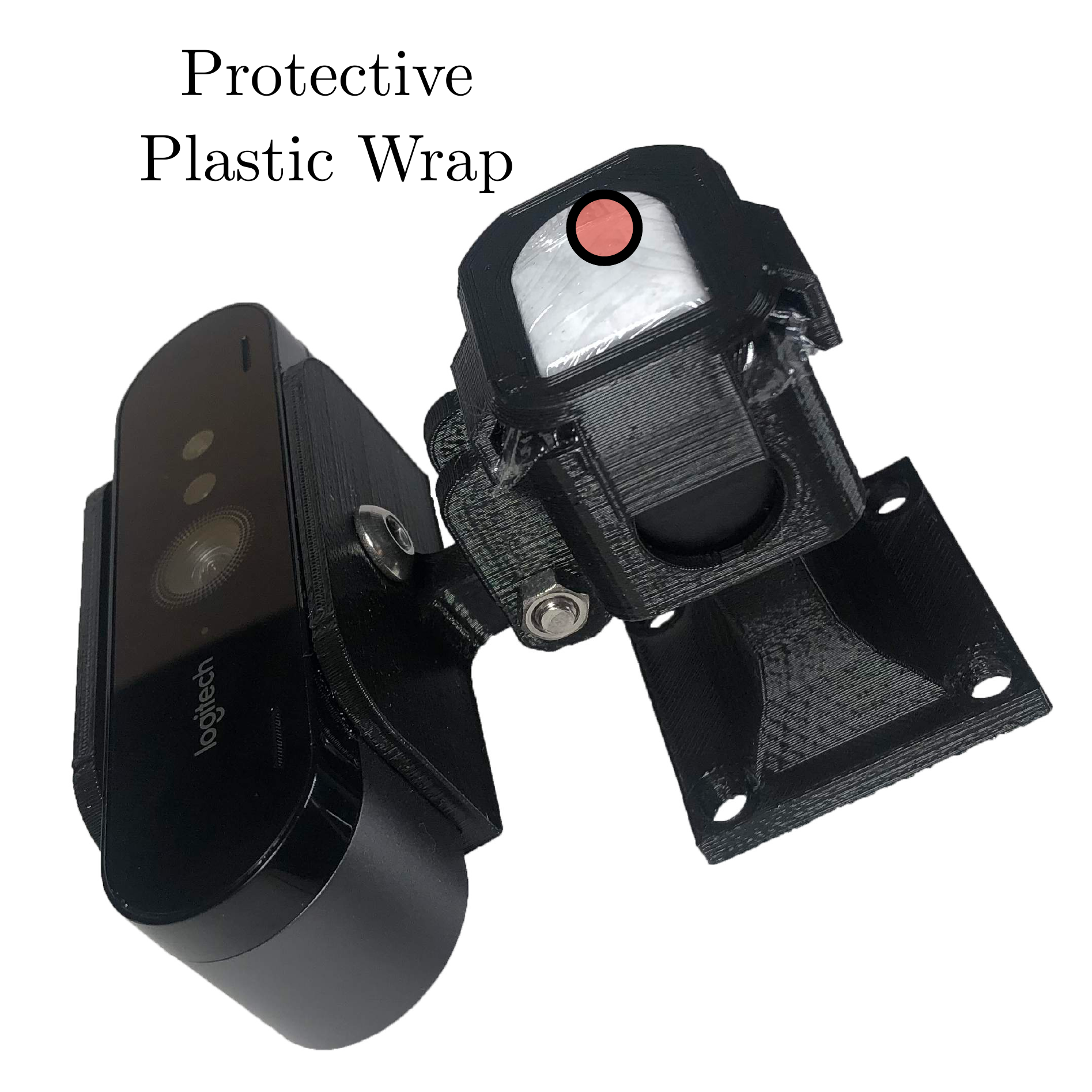}
    }
    \caption{\textbf{Custom Hardware Design:} (a) A CAD model of the mechanical mount designed for data collection with the RGB camera and tactile sensor. (b) The actual sensors and the mount.}
    \label{fig:cad}
    \vspace{-13pt}
\end{figure}

\section{Methods}

\begin{figure*}[t!]
\vspace{1pt}
    \centering
    \subfigure[Anomaly Detection]{
        \label{sfig:anomaly-det}
        \includegraphics[width=0.64\columnwidth]{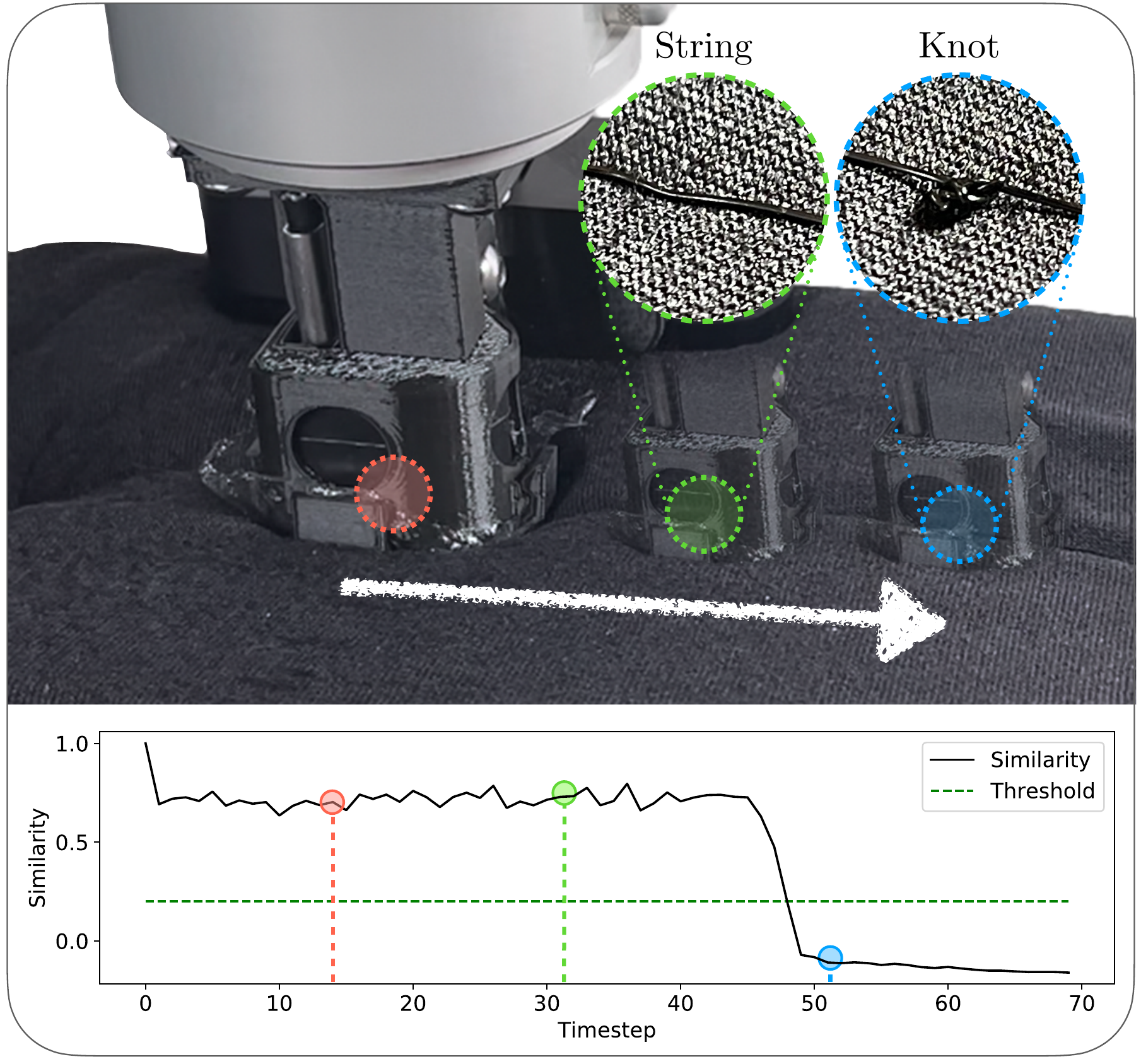}
    }
    \subfigure[Feature Search]{
        \label{sfig:vis-guided}
        \includegraphics[width=0.64\columnwidth]{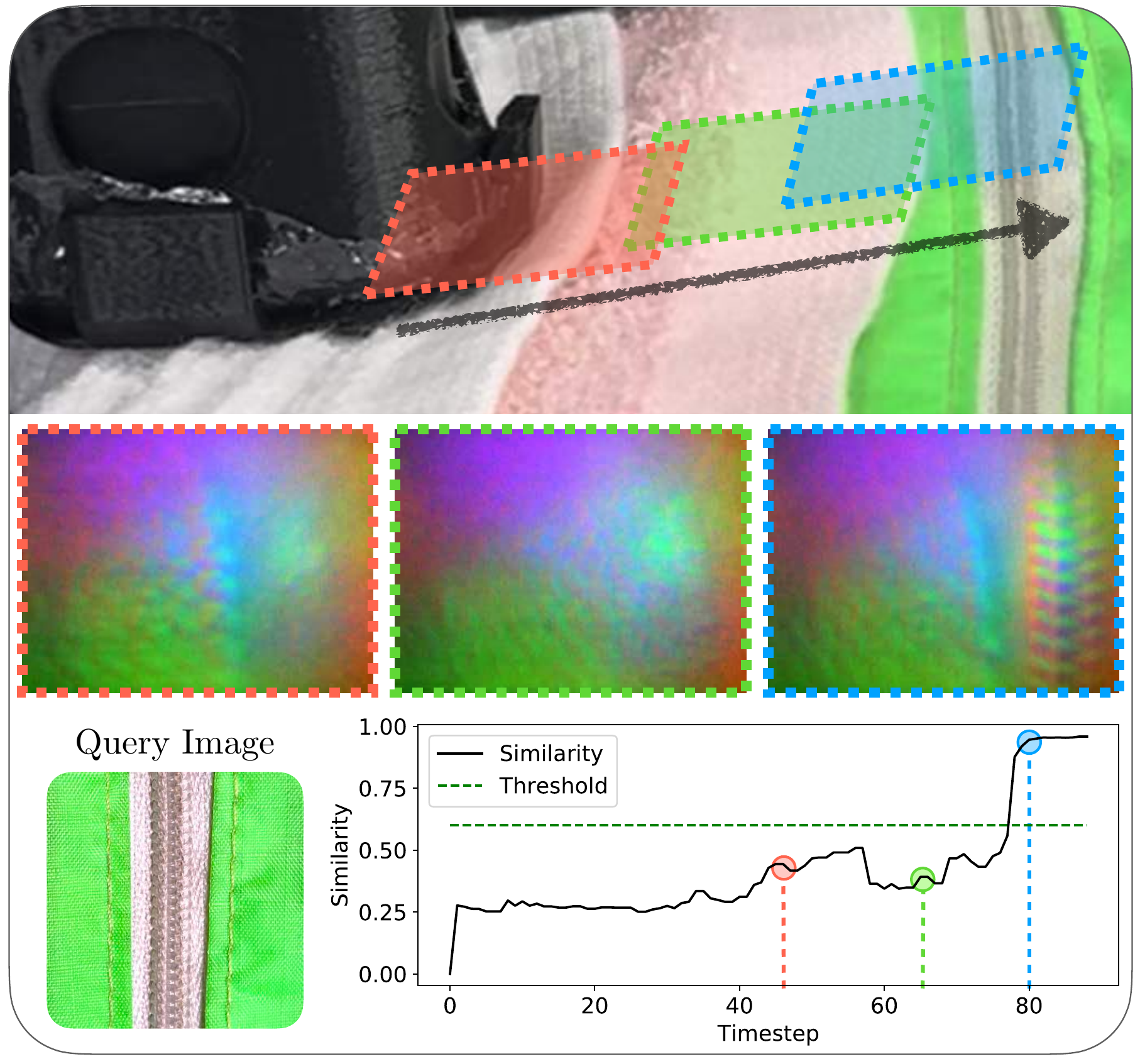}
        \label{sfig:exp-vision}
    }
    \subfigure[Edge Following]{
        \label{sfig:active-servo}
        \includegraphics[width=0.64\columnwidth]{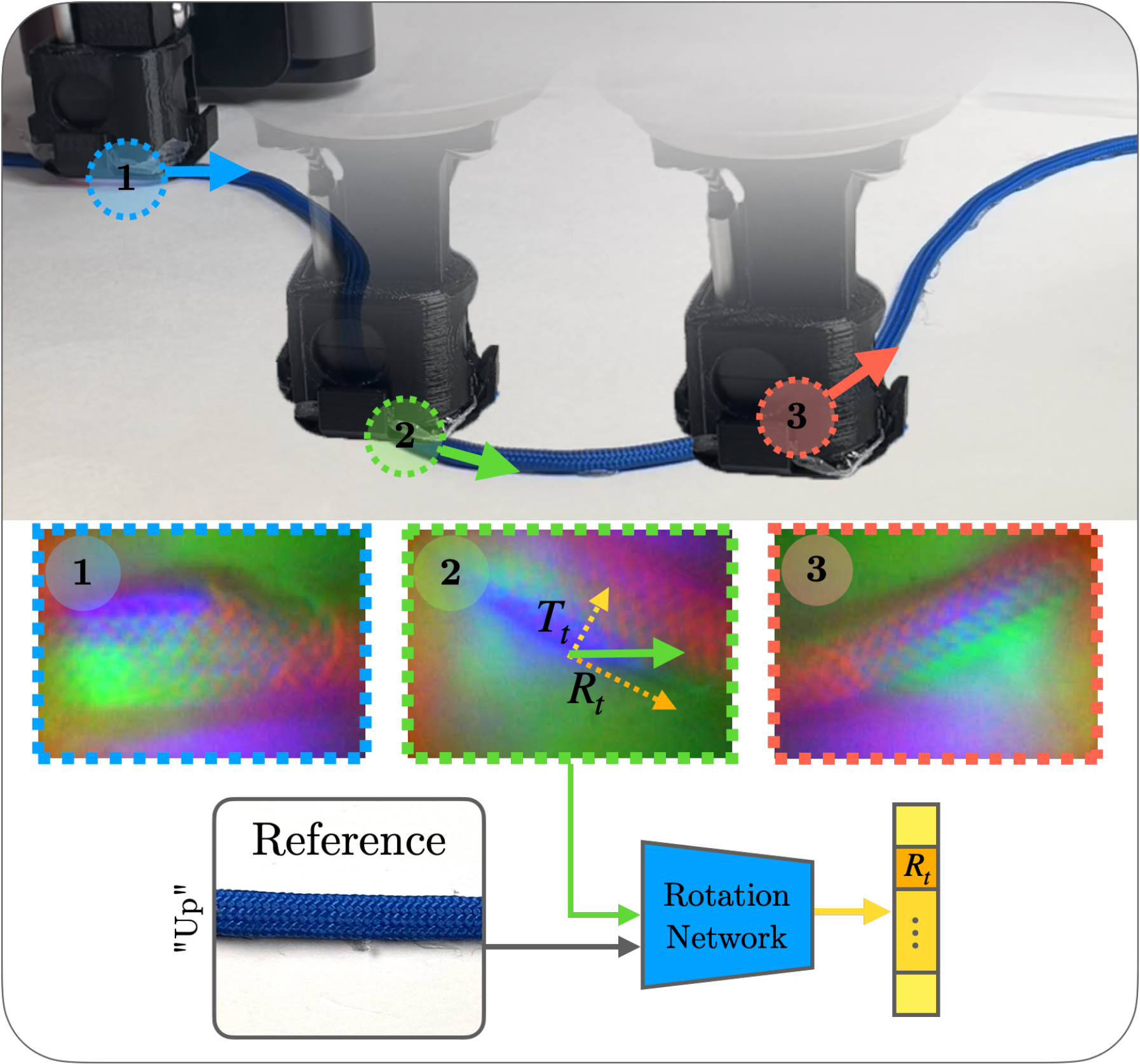}
    }
    \caption{\textbf{Active Sliding Perception Tasks.} (a) Anomaly Detection: The robot uses the learned tactile encoder to detect a knot while sliding over a black thread on a black surface. (b) Feature Search: Here the system is given a query visual image of a zipper and searches the workspace for a tactile reading that is sufficiently similar to the query (i.e., the texture of a zipper). (c) Edge Following: The robot uses rotation predictions to follow a curved cable. Visual inputs are colorized for clarity, the visual network takes in grayscale.}
    \label{fig:exp}
    \vspace{-15pt}
\end{figure*}

\algabbr{} collects self-supervised data using custom hardware (Section~\ref{ssec:data_col}). The collected dataset is then used to train a visuo-tactile latent space (Section \ref{ssec:encoders}) and a rotation prediction network (Section \ref{ssec:rotation}). These networks are used in active sliding perception primitives (Section \ref{ssec:manip_policies}) and passive perception modules (Section \ref{ssec:passive}) for five downstream tasks.

\subsection{Self-Supervised Data Collection}\label{ssec:data_col}

To permit cross-modal representation learning we wish to collect \textit{spatially aligned} pairs of visual and tactile images. To accomplish this, we design a self-supervised data collection pipeline that uses a robot to automatically collect paired images on deformable surfaces defined in Section \ref{sec:problemstatement}. We design a custom end-effector (Figure~\ref{fig:cad}) to hold both sensors parallel to each other with a fixed offset $d$ between the camera and tactile sensor.  Unlike many prior works that train on images with a large field of view, we zoom in closely when collecting visual images so that their field of view is similar to the tactile images, allowing the visual images to show detailed local textures. We capture visual images with a 2$\times$ wider field of view to permit rotation augmentations without border artifacts in the data.

At the beginning of each round of data collection, we normally set up the workspace by laying out 5-15 different objects at stable poses on a flat surface to form a deformable surface. The robot then samples planar translations on a predefined grid with a resolution of \SI{2}cm and a fixed height offset above the surface. For each sample, the system generates random noise uniformly between \SI{-1}cm and \SI{1}cm in translation and between $0^\circ$ and $45^\circ$ in rotation to mitigate aliasing induced by grid sampling. The robot moves the camera above the noised position, takes a visual image using the camera, then translates the constant $d$ offset to align the tactile sensor above the camera observation center. The robot moves down in a stamping motion, stopping when the arm registers a force of 20N. This allows good contact with the sensor without harming the tactile membrane. Thus, the collected tactile and vision images are automatically spatially aligned with a similar field of view. The pipeline takes 11 samples per minute, with minimal human involvement required to periodically reset the surface with new objects. We distribute LED lights close to the workspace to limit self-shadows from the arm. This pipeline is shown in Figure~\ref{fig:data_col}.

\subsection{Latent Space Training}\label{ssec:encoders}
%We decouple the learning of texture and orientation. 
We train a vision encoder $f_\theta: O_\mathrm{vis} \rightarrow \mathcal{Z}$ and a tactile encoder $g_\phi: O_\mathrm{tac} \rightarrow \mathcal{Z}$, where $\mathcal{Z} = \{z \in \mathbb{R}^d: || z|| = 1\}$ is a shared output representation space with $d \ll H \times W \times 3$. We use the collected data (Section~\ref{ssec:data_col}) and contrastive learning to learn a visuo-tactile association in a latent space. We choose a contrastive loss because of its prior success in embedding different modalities within the same space~\cite{Radford2021LearningTV,yang2022touch}.
We use grayscale camera images as input to the visual encoder to avoid overfitting to colors of materials,
and we augment all images with flips and color jitter to make the network less sensitive to input image orientation and lighting. For each visuo-tactile image pair we also apply random rotation uniformly between $0^\circ$ and $360^\circ$ to encourage rotation-invariant learning.
During training we use InfoNCE loss on the pairwise dot products of the embedding vectors, which intuitively pushes the correct pair of visuo-tactile readings together and all others apart. We constrain embedding vectors to have unit length, as in CLIP \cite{Radford2021LearningTV}. Mathematically, this loss is 
$L = \text{cross\_entropy}(I_e\times T_e^\text{T}, [0...B])\,,$
where $T_e$ and $I_e$ are batched embeddings of the tactile and visual images respectively and $B$ is the batch size. The outputs of the trained encoders are $d$-dimensional embeddings in the learned latent space. See Appendix \ref{app-encoder} for more details on parameters and augmentations.

\subsection{Rotation Prediction Network}\label{ssec:rotation}
We train a separate rotation network $h_\psi: O_\mathrm{vis} \times O_\mathrm{tac} \rightarrow [0, 1]^{N_b}$ on the same dataset to learn the difference in rotation between a vision image and a tactile image. We use color augmentation to prevent potential spurious lighting statistics in the data \cite{Efros2015}. We discretize the rotations into $N_b$ buckets uniformly between 0$^\circ$ and 360$^\circ$ and formulate the problem as classification where the input is a concatenated visuo-tactile image pair and the output is a distribution over bucket indices. Classification aids in learning multi-modality in rotations ($\pm 180^\circ$). See Appendix \ref{app-rotation} for further details.

\subsection{Sliding Perception Primitives}\label{ssec:manip_policies}
We implement three active sliding perception primitives for use in downstream tasks on garment features (Figure \ref{fig:exp}) using the trained encoders and rotation network without fine-tuning. The first two focus on \textit{locating} features of interest, and the third on \textit{following} these physically with the sensor.

\subsubsection{Anomaly Detection}
To inspect garments, humans often slide their hands across the surface to detect textures which stand out from the background. We mimic this behavior by sliding the tactile sensor across a deformable surface and using the trained tactile encoder to localize tactile anomalies in materials. The system takes tactile readings while sliding and encodes them with $g_\phi$. The system maintains a sliding window $\mathcal{B}_h$ containing past tactile embeddings and continually evaluates the median cosine similarity between the current embedding and the embeddings in the window. If the median similarity crosses a predefined threshold $m_1$, the system outputs that it has detected an anomaly candidate and slows the robot down to take $n_1$ more readings. If the median similarity remains below $m_1$, the system outputs an anomaly; otherwise, the system resumes sliding at the original speed.
 
\subsubsection{Feature Search}
When searching for a specific tactile feature in a garment (e.g., a button in a shirt), anomaly detection is insufficient due to the existence of distractors (e.g. seams). 
To remedy this, we search for a tactile reading which matches 
a \textit{visual} query image of the target. This visual image is embedded into the shared latent space. While sliding, the system computes the cosine similarity (dot product) between the current tactile embedding and the query image embedding. If the similarity is above $m_2$, the policy slows down the robot to takes $n_2$ more tactile readings. If the median similarity remains greater than $m_2$, the system outputs that it has found the target; otherwise, it resumes sliding at the original speed. 

\subsubsection{Edge Following}
\label{sec:method_servoing}
Another useful primitive for tactile manipulation is actively servoing along garment features like edges, cables, and seams. Prior approaches utilize analytic methods \cite{She2020CableMW} or supervised learning with human labels \cite{alberto2022,servoing1}. In this work, the robot servos along material features specified purely from visual images, leveraging the visuo-tactile pretraining to avoid the need for a servoing dataset. Given an input visual image specifying the feature to follow as well as the direction to follow it in, the system uses the rotation network to predict the relative rotation between the query visual image and the current tactile image. This rotation provides a translation in the tactile frame, $R_t$, to move in which follows the feature. Sometimes the garment feature may drift orthogonally to $R_t$, so we also estimate a feature-centering offset. To do this, we approximate network confidence as the variance of the predicted rotation distribution. We translate the input image orthogonally to $R_t$ in both directions by half the image width, and choose the offset as the translation which maximizes confidence. Maximizing confidence serves as a centering signal because higher confidence corresponds to views with the feature more visible.

\begin{figure}[t!]
\vspace{1pt}
    \centering
    \includegraphics[width=0.97\columnwidth]{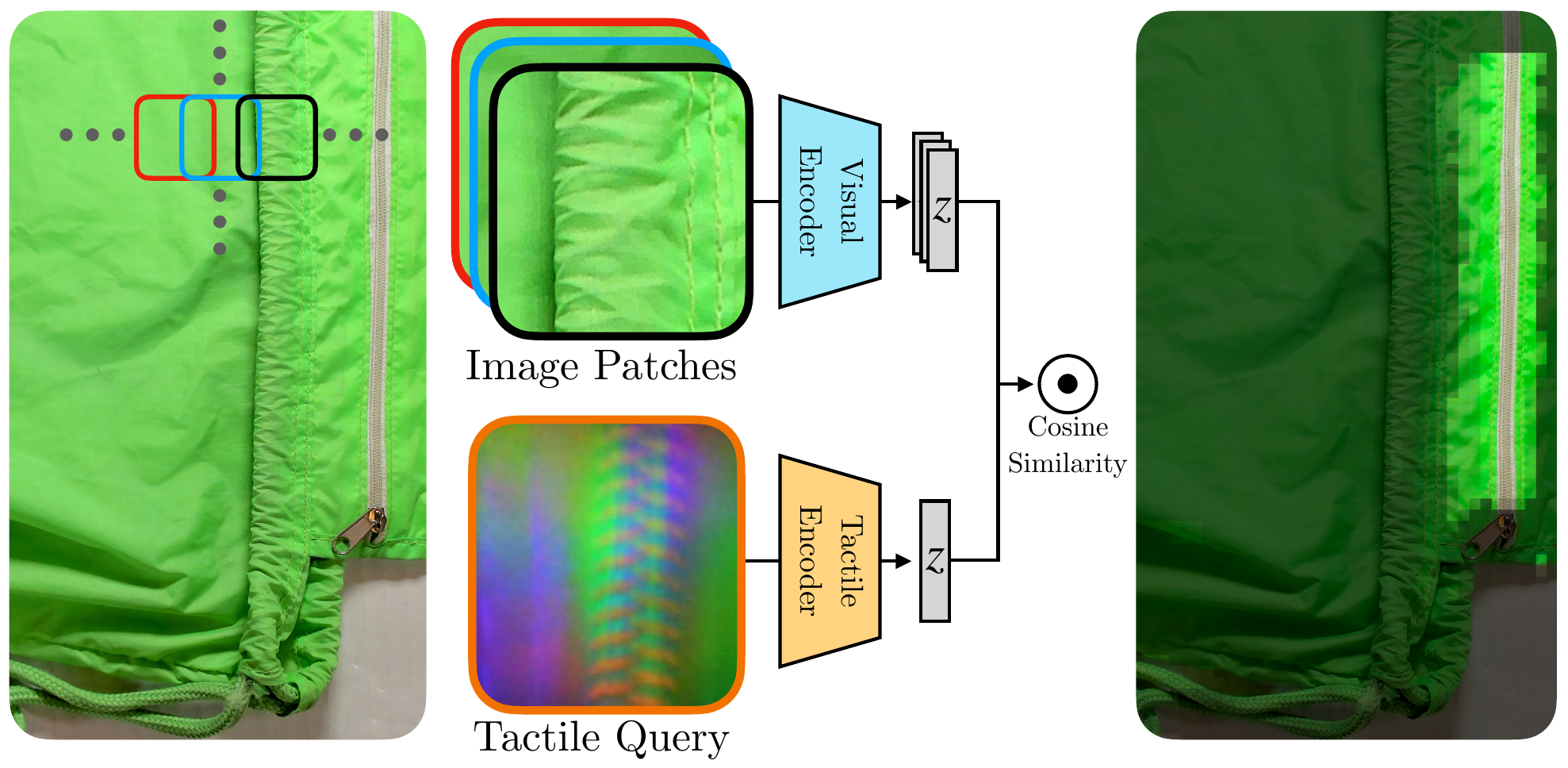}
        \caption{\textbf{Tactile Localization:} An input tactile image of the zipper is compared to discreteized patches from a visual image of the entire scene. In this example, the heatmap shows high probability of a match near the zipper. Note that visual images are colorized for clarity, the visual encoder takes in grayscale.}
    \label{fig:contact_loc}
    \vspace{-15pt}
\end{figure}

\subsection{Passive Perception Modules}\label{ssec:passive}
We develop two modules that use the pretrained, frozen encoders for passive tactile perception: a localizer module and a classifier module.
\subsubsection{Contact Localizer}
Given a top-down visual image of the entire workspace (a sort of ``visual map") and a tactile image, this module outputs a probability distribution over possible contact locations in the visual image. We divide the input visual image into a grid with uniform patch size and resolution and compute cosine similarity of each patch embedding with the tactile embedding as shown in Figure~\ref{fig:contact_loc}. This produces a distribution over patch locations of the predicted similarity between vision and touch. 
\subsubsection{Feature Classifier}
\label{sec:method_classifier}
\begin{figure}[t!]
    \centering
    \includegraphics[width=0.94\columnwidth]{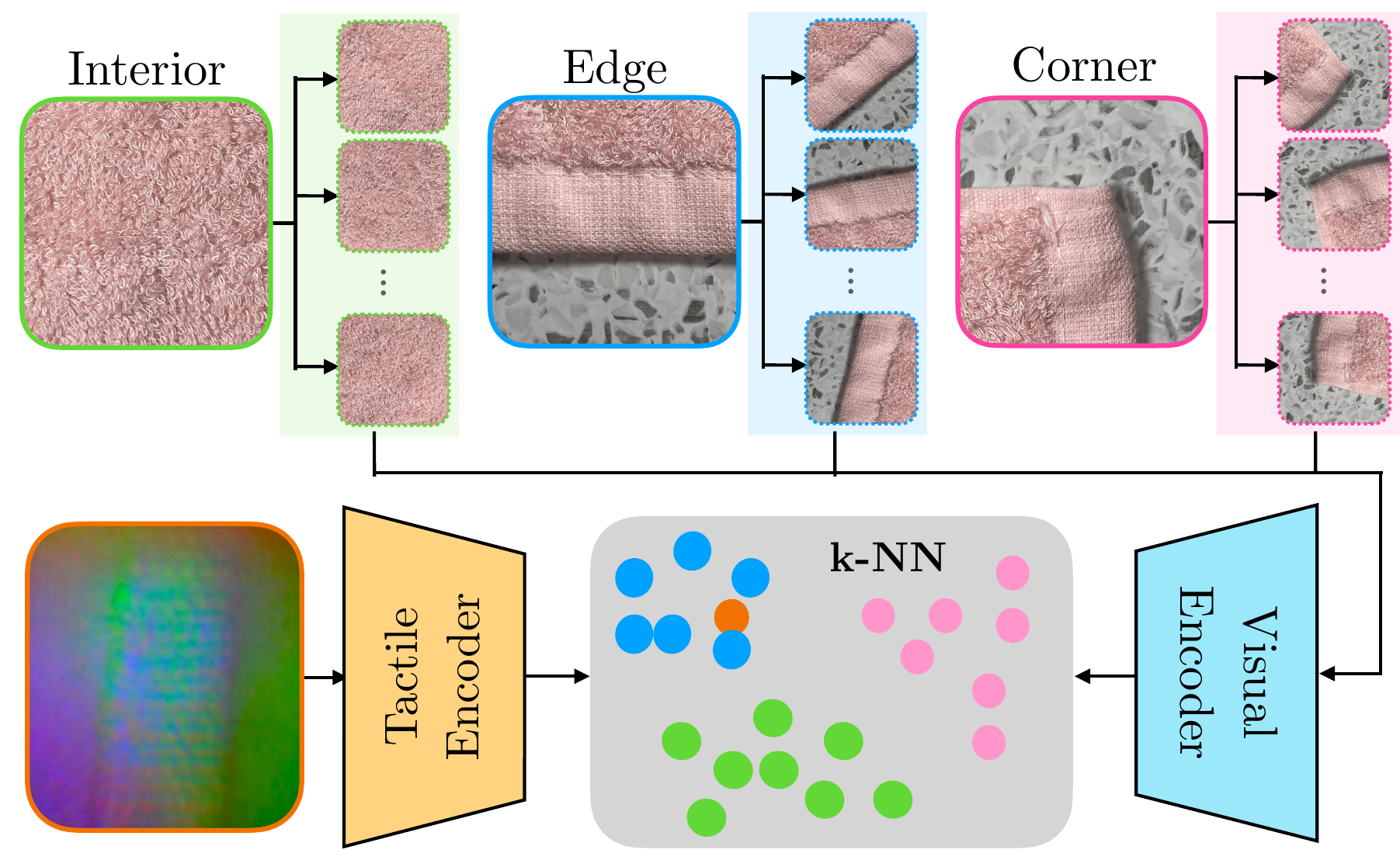}
    \caption{\textbf{Tactile Classification:} Given canonical visual images of categories (images in solid line), we augment these inputs (images in dashed line) and embed them into the latent space, then apply k-NN classification on tactile input images. Images are colorized for clarity.}
    \label{fig:classfier}
    % \vspace{-15pt}
\end{figure}

Classification of features in input tactile images (e.g., distinguishing between towel edges and corners) is a common task. Prior work performs tactile classification with supervised learning \cite{alberto2022}, but human supervision is time-consuming. Instead, we construct a classifier from trained encoders by providing canonical visual images of the classes we consider, which can be rapidly captured. We augment these visual images with the same augmentation as performed during training and embed them into $\mathcal{Z}$. We use the weighted \textit{k}-nearest neighbors algorithm to classify the tactile input (Figure~\ref{fig:classfier}), a technique shown to be effective in classification with latent spaces \cite{yuinstanceseg} in computer vision.
% \vspace{-0.1cm}

\section{Physical Experiments}\label{exp}
We assess the utility of \algabbr{} through performance on a diverse range of tasks centered around garment feature localization and following. Throughout all experiments, the same pretrained networks are used without fine-tuning, to evaluate the flexibility of the representations learned from visuo-tactile pretraining.

We collect self-supervised data (Section~\ref{ssec:data_col}) with a UR5 robot arm, a DIGIT tactile sensor~\citep{Lambeta_2020}, and a Logitech BRIO webcam. This same setup is used for physical experiments.
The data consist of 4500 visuo-tactile pairs over 10 deformable surface environments. It took 7 hours of unsupervised robot time to collect all data samples, which involved about 20 minutes of human time to reset the environment with new objects between sampling.

Both tactile and visual networks use an input resolution of $128\times128$, to which all input images are resized to.  We train a texture latent space (Section~\ref{ssec:encoders}) with an embedding dimension of $d = 8$.
 Visual and tactile encoders use a RegNetY$\_$800MF backbone \cite{He2016DeepRL} with an altered final layer because of its performance per MFLOP. The rotation network uses a RegNetY$\_$800MF backbone with an altered first layer to support 2 stacked images, and a final layer to output the rotation bins. We initialize all networks with weights pretrained on ImageNet~\cite{ILSVRC15} classification and train all parameters. See the Appendix for more training and parameter details.

\subsection{Active Sliding Tasks}
 For these tasks, the experiment starts with the robot pushing down at the given start position until a force of \SI{20}N is reached, to establish good tactile contact. The robot slides at a speed of \SI{1}cm/s acquiring tactile readings at a rate of \SI{10}hz. We do not use closed-loop control on the force applied, since using the UR5's position control is sufficient to maintain consistent contact on the flat surfaces we test on. All the experiments below are conducted without vision input other than the query or reference visual images. The tactile sensor mount includes a thin layer of protective plastic wrap over the sensor which aids sliding primitives by mitigating friction.
\subsubsection{Anomaly Detection}
For this experiment we set $|\mathcal{B}_h|=40$, $m_1=0.3,$ and $ n_1=20$. The robot is slowed to 20\% speed when an anomaly candidate is detected. A trial is considered successful if the robot stops at a tactile reading that contains the anomaly. Each experiment consists of 10 trials with varying start locations. Figure~\ref{sfig:anomaly-det} shows an example.
As a baseline, we apply the same buffer and active confirmation technique but using a threshold on the L2 pixel distance between tactile images over time which detects sharp changes in the appearance of the tactile image. 

We consider the following 2 settings:
\textbf{Knot detection in string}:
 We fix a \SI{1}{\milli\metre} black surgical thread on top of black cloth and tie a knot anomaly to detect. We conduct experiments on a \SI{1}{\milli\metre} knot and a \SI{3}{\milli\metre} knot respectively.
\begin{table}[t!]
    \centering

    \caption{Anomaly detection success rate.}
    \label{tab:sliding}
    \begin{tabular}{lccc}
        \toprule
          Algorithm & Knot 3mm & Knot 1mm & Lump \\
        \midrule
         Image diff & $\mathbf{9/10}$ & $0/10$ & $2/10$ \\
         \algabbr & $\mathbf{9/10}$ & $\mathbf{8/10}$ & $\mathbf{5/10}$ \\
         \bottomrule\\
    \end{tabular}

    \caption{Feature search success rate.}
    \label{tab:vgs}
    \begin{tabular}{lccc}
        \toprule
          Algorithm & Zipper & Garment & Button \\
          \midrule
         Anomaly Det. & $0/10$ & $1/10$ & $5/10$ \\
         \algabbr{}& $\mathbf{10/10}$ & $\mathbf{10/10}$ & $\mathbf{10/10}$ \\
         \bottomrule\\
    \end{tabular}

    \caption{Edge following distance \% completed.}
    \label{tab:servo}
    \begin{tabular}{lccc}
        \toprule
         Algorithm & S-Shaped Cable & Dress Seam & Shirt Seam \\
          \midrule
         Ellipse Fit & $12.7 \pm 24.0$ & $43.1 \pm 35.4$ & $18.8 \pm 9.6$ \\
         \algabbr & $\mathbf{78.7 \pm 3.4}$ & $\mathbf{92.2 \pm 23.3}$ & $\mathbf{92.4 \pm 19.1}$ \\
         \bottomrule
    \end{tabular}
    \label{tab:sliding-anom-detec}
    \vspace{-5pt}
\end{table}
\textbf{Lump defect detection}:
To simulate a manufacturing defect in a piece of soft foam, we embed a \SI{3}mm hex nut in foam, \SI{1}cm beneath the surface. %The system detects the nut in 8/10 trials. 
\begin{table}[t!]
\vspace{2pt}
    \centering
    \caption{Classification results on towel edges, corners, and interiors. We use K-nearest neighbor classification on the latent embeddings of a small set of canonical \textit{visual} images to classify new \textit{tactile} images (Sec.~\ref{sec:method_classifier}). N refers to the number of datapoints, and ``primary confusor" is the mode of mis-classifications.}
    \begin{tabular}{cccc}
    \toprule
         \textbf{Category} & \textbf{Accuracy (\%)} & \textbf{Primary Confusor} & \textbf{N} \\ 
         \midrule
         Edge & 75 & Corner&36\\
         Corner & 93 & Edge&30\\
         Interior & 74 & Edge&43\\
         \hline
         Overall & 80 & -- &109
    \end{tabular}
    \label{tab:classification}
    \vspace{-17pt}
\end{table}
Results (Table~\ref{tab:sliding}) show that \algabbr{} achieves an average 73\% success rate on the three settings, while L2 distance achieves an average 37\% success rate. The baseline's performance is appreciable for prominent features like a large knot, but it is insufficiently sensitive for more subtle features like the smaller knot. Both techniques achieve lower success rates on lump detection due to the more subtle changes in the tactile readings, though \algabbr{} is more robust.
\subsubsection{Feature Search}
In these experiments, to test if \algabbr{} can differentiate targets from multiple distractors, we include false tactile anomalies along the path of the robot to the correct feature. We use $m_2=0.6$ and $n_2=60$. The robot is slowed to 20\% speed when a target candidate is detected. Figure~\ref{sfig:vis-guided} shows an example. We use Anomaly Detection as a baseline to test if the distractions are indeed distractors, and run 10 trials for each experiment varying the start location. An experiment is a success if the robot stops with the feature specified by the visual image in the tactile image. We consider the following three settings:

\textbf{Multi-texture garment}: The robot searches for a ruffled portion of a dress amidst seams and crumples, a subtle difference in tactile texture to detect.
\textbf{Finding a zipper among multiple edges}:

The robot searches for a zipper on a multi-textured surface.
We layer two different towels on a bag with a zipper, and the query image is a visual image of the zipper (Figure~\ref{sfig:vis-guided}). 

\textbf{Finding a button in a blouse}: The robot slides on a blouse until a button is found, ignoring seams along the way. 

The results, summarized in Table~\ref{tab:vgs}, show that \algabbr{} achieves a 100\% success rate while the baseline achieves 20\% average success rate, suggesting that the learned latent space can successfully correlate tactile and vision information, while anomaly detection easily mis-fires on distractor features.

\subsubsection{Edge Following}
For each experiment, we provide a visual image of a target feature and specify the direction to follow in that image. We smooth the direction vector ${R}_{t}$ obtained from the rotation network by $\hat{R}_{t} = \lambda R_{t} + (1-\lambda) R_{t-1}$, where $\lambda =0.8$ is chosen empirically. This results in the overall sliding direction of $\hat{R}_{t} + T_{t}$ (see Sec \ref{sec:method_servoing} for details). We measure the distance successfully traveled without losing contact. As a baseline, we use the contact ellipse estimation from PyTouch~\cite{pytouch} and move along the major axis. We average over 10 trials for each test object. We conduct the experiments on a \SI{2}mm diameter, \SI{30}cm length \textbf{S-shaped cable} fixed on the work surface, a \textbf{dress seam} in the middle of a dress with wrinkles on the side and a \textbf{shirt seam} near a shirt collar.

The percent distance traveled is in Table \ref{tab:servo}. While the baseline travels up to 43.1\% of the total distance, our policy travels at least 78.7\% of the total distance. The baseline is sensitive to flat textures that confuse the ellipse fitting, while our rotation network prediction is robust to local textures. The major failure cases of our method are ambiguity in servoing direction ($\pm 180^\circ$) and errors from the centering offset $T_t$ if the feature drifts too much. This effect is especially visible in the S-shaped cable, where the direction changes drastically during a rollout, whereas the seams only have a gently curve.
\begin{figure}[t!]
\vspace{1pt}
    \centering
    \includegraphics[width=1.0\columnwidth]{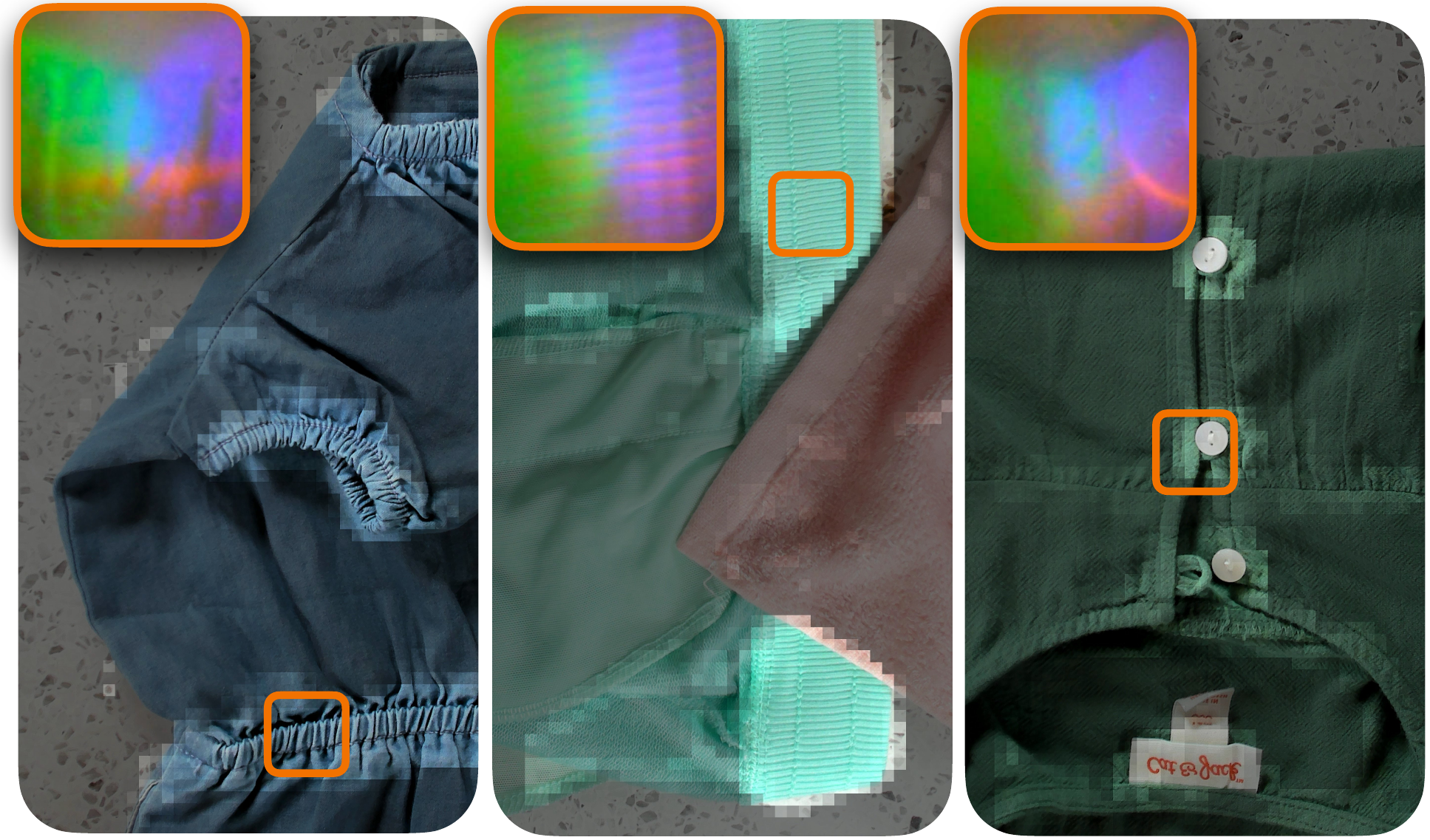}
    \caption{\textbf{Localization Visualizations:} We show 3 illustrative contact localization results; a wrinkled dress seam, an elastic band, and blouse buttons. The heatmaps are composited onto the original image by scaling them between $[0,1]$, applying gamma correction $\gamma=1.6$, and multiplying them by the value channel. Orange boxes denote the location of the tactile query.}
    \label{fig:heatmap_res}
    \vspace{-20pt}
\end{figure}
\subsection{Passive Perception}
\subsubsection{Contact Localization}
\label{exp:localization}
Since many locations in the visual image may generate nearly identical tactile readings, such as the interior of a fabric, we provide qualitative visualizations of localization heatmaps in Figure~\ref{fig:heatmap_res}. Similar textures as the ground truth touch location are smoothly activated, suggesting the encoder has learned a mapping between visual and tactile appearance rather than overfitting to specific locations.

\subsubsection{Feature Classification}

We manually collect a dataset of 109 tactile images of corners, edges, and interiors of 2 different towels, and use 4 canonical visual images per category to instantiate the K-NN classifier. We use 100 augmentations of the 4 canonical images to smooth decision boundaries and k=50. Results are shown in Table \ref{tab:classification}. We hypothesize that the edges are occasionally confused as corners because they look locally similar, while the interiors are misclassified as edges due to the similarity in texture between the interior of one towel to the edge texture of the other.

\section{Limitations}
This approach has notable limitations. First, the similarity scores obtained via the latent space can be difficult to interpret, making performance on query tasks sensitive to a chosen threshold. Techniques to avoid thresholding could incorporate probabilistic estimation methods like particle filters to account for uncertainty, analogous to \cite{midastouch}. In addition, the deformable surfaces were largely planar. Some human time is still required to reset the workspace between rounds of data collection; fully automating the data collection procedure could improve scalability. In this work the test garments were contained within the self-supervised dataset, so the generalization performance remains to be studied at scale. Another consideration is the size of the cropped visual query images: during usage the input visual images must be within the training distribution or results can become unreliable. In this work the scale was manually chosen, but in the future the system could select this scale based on depth readings from an RGBD camera, or automatically based on the similarity score.

\section{Conclusion}
In this paper, we propose Self-Supervised Visuo-Tactile Pretraining: a self-supervised framework for learning visuo-tactile representations. We apply the learned representations to 3 active and 2 passive perception tasks on garment features. Results across this diverse range of tasks suggest the flexibility of the learned representations for diverse tasks including contact localization, tactile feature search, feature classification, anomaly detection, and edge following. While we focus our attention on garment features in this work, SSVTP is not limited to this application and poses a step toward task-agnostic visuo-tactile representation learning for robot control. 
\section*{Acknowledgments}
This material is based upon work supported by the National Science Foundation Graduate Research
Fellowship Program under Grant No. DGE 2146752. Any opinions, findings, and conclusions
or recommendations expressed in this material are those of the author(s) and do not necessarily reflect the views of the National Science Foundation.
This research was performed at the AUTOLAB at UC Berkeley in affiliation with the Berkeley AI Research (BAIR) Lab, and the CITRIS ``People and Robots" (CPAR) Initiative.
Funded by the German Research Foundation (DFG, Deutsche Forschungsgemeinschaft) as part of Germany’s Excellence Strategy – EXC 2050/1 – Project ID 390696704 – Cluster of Excellence “Centre for Tactile Internet with Human-in-the-Loop” (CeTI) of Technische Universität Dresden.

\bibliographystyle{plainnat}
\bibliography{references}

\section{Appendix}
\label{appendix}
\subsection{Encoder training details}
We choose the encoder dimension $d=8$ via grid search over $d \in [8, 16, 32, 64]$, and selected based on qualitative performance on a held-out test set of contact localization and classification. Though it may seem surprising that a lower dimension performs better, this is because given a relatively small dataset (4500 pairs), a lower dimension latent space has a regularizing effect preventing overfitting. We optimize using Adam~\cite{Kingma2015AdamAM} with default parameters, and exponential learning rate decay with $\gamma=0.95$, gradient clipping at $0.5$, and a batch size of $\text{B}=512$. This batch size was the largest that could fit in our GPU memory, following literature from the vision-language supporting large batch sizes as key for latent spaces\cite{Radford2021LearningTV}.
\label{app-encoder}
\subsubsection{Augmentation}
For training the latent space, we apply color augmentations to both tactile and visual images. Though visual images are grayscale, some of these still affect the appearance like contrast and brightness. For tactile images, these we use PyTorch's ColorJitter with parameters {\tt brightness=(0.9,1.1)}, {\tt contrast=(.9,1.1)}, {\tt saturation=0.2}, {\tt hue=0.05}. For visual images, these are {\tt brightness=(0.8,1.1), contrast=(.7,1.3), saturation=0.2}
We also apply spatial augmentations \textit{independently} to tactile and visual images. These are rotation augmentation ($\pm360^\circ$), translation (2\% of image scale), and horizontal flips. These augmentations are applied randomly during dataloading; as such there is no fixed dataset size used, just procedurally generated augmentations of the original 4500 pairs.
\subsection{Rotation network training}
We train the rotation network (Section~\ref{ssec:rotation}) with a batch size of $\text{B}=32$, $N_b=18$, and Cross Entropy Loss. 
\subsubsection{Augmentation}
During rotation network training we use the same color augmentations as in the latent space encoder, and apply rotation augmentations to the visuo-tactile pairs \textit{non-independently}, so that the rotation correlation between pairs is preserved. Then, we randomly select a \textit{relative} rotation to transform the tactile image, thus obtaining a training datapoint with a labeled, random rotation. This rotation is converted to the bin index for the label.

\label{app-rotation}

\end{document}